\documentclass[letterpaper, 10 pt, conference]{ieeeconf}  

\IEEEoverridecommandlockouts                              

\usepackage{xcolor}
\usepackage{hyperref}
\usepackage{graphicx}
\usepackage{bm}
\usepackage{dsfont}
\usepackage{booktabs}
\usepackage{multirow}
\usepackage{amsmath}
\usepackage{amssymb}
\usepackage{amsfonts}
\usepackage{caption}
\usepackage{subcaption}
\usepackage[noadjust]{cite}
\usepackage{soul}

\makeatletter
\newcommand\footnoteref[1]{\protected@xdef\@thefnmark{\ref{#1}}\@footnotemark}
\makeatother

\title{\LARGE \bf
Competency-Aware Planning for Probabilistically Safe Navigation Under Perception Uncertainty
}

\author{Sara Pohland$^{1}$ and Claire Tomlin$^{2}$
\thanks{$^{1}$Department of EECS,
        UC Berkeley
        {\tt\small spohland@berkeley.edu}}%
\thanks{$^{2}$Department of EECS,
        UC Berkeley
        {\tt\small tomlin@berkeley.edu}}%
}

\begin{document}

\maketitle
\thispagestyle{empty}
\pagestyle{empty}

\begin{abstract}

Perception-based navigation systems are useful for unmanned ground vehicle (UGV) navigation in complex terrains, where traditional depth-based navigation schemes are insufficient. However, these data-driven methods are highly dependent on their training data and can fail in surprising and dramatic ways with little warning. To ensure the safety of the vehicle and the surrounding environment, it is imperative that the navigation system is able to recognize the predictive uncertainty of the perception model and respond safely and effectively in the face of uncertainty. In an effort to enable safe navigation under perception uncertainty, we develop a probabilistic and reconstruction-based competency estimation (PaRCE) method to estimate the model's level of familiarity with an input image as a whole and with specific regions in the image. We find that the overall competency score can accurately predict correctly classified, misclassified, and out-of-distribution (OOD) samples. We also confirm that the regional competency maps can accurately distinguish between familiar and unfamiliar regions across images. We then use this competency information to develop a planning and control scheme that enables effective navigation while maintaining a low probability of error. We find that the competency-aware scheme greatly reduces the number of collisions with unfamiliar obstacles, compared to a baseline controller with no competency awareness. Furthermore, the regional competency information is particularly valuable in enabling efficient navigation.

\end{abstract}

\section{INTRODUCTION}

Within the area of unmanned ground vehicle (UGV) navigation in unstructured environments, learning-based components have become commonplace and seemingly necessary \cite{guastella_learning-based_2020}. Deep neural network (DNN)-based perception models have proven to be particularly useful for accurately analyzing the traversability of the terrain, preventing collisions, and enabling effective navigation \cite{guastella_learning-based_2020}. Unfortunately, as data-driven and black-box methods, DNN-based perception models face inherent limitations: they lack transparency and explainability \cite{saleem_explaining_2022}, are often overconfident in their predictions \cite{guo}, are sensitive to shifts in the input data distribution \cite{ovadia}, and can fail quite surprisingly and ungracefully \cite{Nguyen_2015_CVPR}.

\begin{figure}[t!]
    \centering
    \captionsetup{width=\columnwidth}
    \includegraphics[width=\columnwidth]{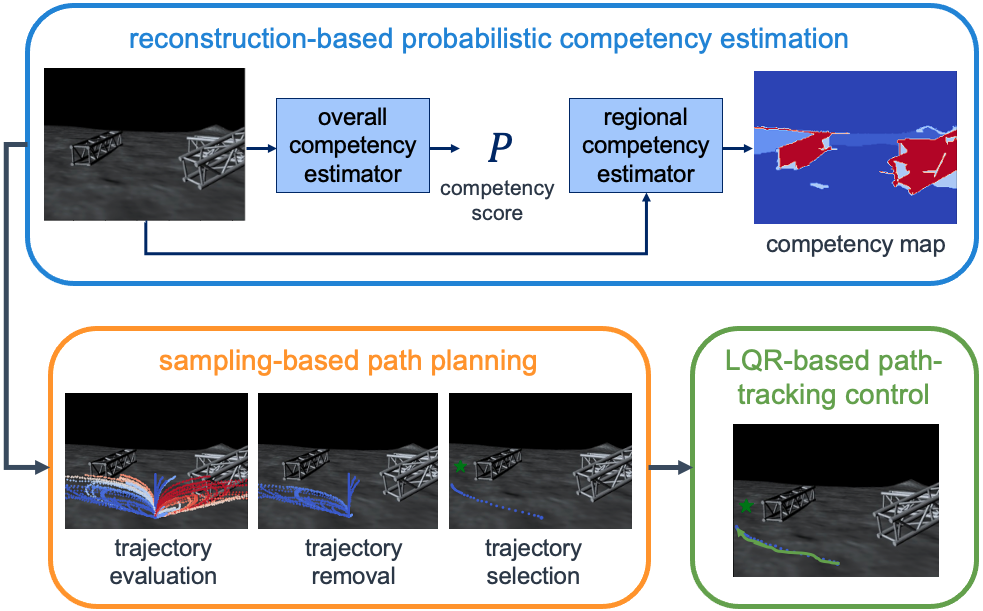}
    \caption{We develop a probabilistic reconstruction-based method for estimating the overall and regional competency of a perception model for a given input image. This competency score and competency map are used to design a sampling-based path planning algorithm that evaluates the minimum competency associated with sampled paths, removes trajectories that fall below the minimum probability threshold, and selects the best path towards the goal. An LQR-based controller is then used to track the planned path.}
    \label{fig:overview}
\end{figure}

With the prevalence of DNN models in many real-world applications and concern about their failure modes, there has been extensive research on quantifying uncertainty in these models \cite{gawlikowski_survey_2023} and detecting inputs that are outside of their training distribution \cite{yang_generalized_2022}. This has enabled the development of uncertainty-aware navigation schemes that explicitly consider the existence of uncertainty arising from DNN-based perception models. However, existing navigation schemes generally rely on uncertainty estimation methods that do not perform well for data outside of the model's training distribution \cite{schwaiger_is_nodate} or on methods that provide binary decisions about whether an input is anomalous based on the training set \cite{quinonero-candela_dataset_2009}. These methods do not allow for the design of probabilistically safe control schemes that maintain performance under multiple aspects of predictive model uncertainty. 

To address these limitations, we develop a competency-aware perception, planning, and control framework to enable safe navigation in complex environments under DNN-based perception model uncertainty (Figure \ref{fig:overview}). In particular, (1)~we develop an overall competency score that represents the probability that the model's prediction is correct for a given image. (2)~We propose a regional competency method that estimates the probability that  regions in the environment are unfamiliar to the model. (3)~We generate a method for evaluating the probability of error associated with a proposed control sequence based on the competency estimates. (4)~We develop a navigation scheme that maintains efficient performance, while ensuring the probability of error remains below a user-defined error threshold. (5)~We evaluate this control scheme in a challenging lunar environment with unfamiliar obstacles and demonstrate the ability of our navigation framework to successfully and efficiently navigate to goal regions, while maintaining a low probability of error.

\section{BACKGROUND \& RELATED WORK}

Field robots used in search and rescue, planetary exploration, and agricultural tasks must navigate in unstructured environments, which lack clearly viable paths or helpful landmarks \cite{guastella_learning-based_2020}. In such environments, perception provides the necessary information to make the vehicle aware of the surrounding environment. Within this context, most work has focused on DNN-based methods to analyze the capability of a UGV to stably reach a terrain region and determine appropriate control actions \cite{guastella_learning-based_2020}. Without properly considering the uncertainty surrounding the predictions of these DNN-based perception models, vehicles that rely on these models can face  unsafe situations and system failures. This has led to a branch of uncertainty-aware navigation focused on dealing with perception model prediction uncertainty.

\subsection{Uncertainty Quantification (UQ)} \label{sec:uncertainty}

Predictive uncertainty generally encapsulates both data/aleatoric uncertainty, which arises from complexities of the data (i.e., noise, class overlap, etc.), and model/epistemic uncertainty, which reflects the ability of the perception model to capture the true underlying model of the data \cite{yarin_gal_uncertainty_2016}.  The modeling of these uncertainties can generally be divided into methods based on (1)~Bayesian neural networks (BNNs) that extract uncertainty as a statistical measure over the output of a BNN \cite{neal-1992,neal_bayesian_1996}, (2)~deterministic neural networks (NNs), often relying on Monte Carlo (MC) dropout as approximate Bayesian inference \cite{dropout}, and (3)~ensembles of NNs that combine the predictions of multiple deterministic networks to form a probability density function \cite{lakshminarayanan_simple_2017}. While these methods address the typical overconfidence in NN predictions with better-calibrated confidence scores, they generally focus on predictions for in-distribution inputs, which come from the same distribution as the training data. These approaches are not sufficient in general to appropriately assign confidence scores for out-of-distribution (OOD) inputs that differ significantly from those seen during training \cite{schwaiger_is_nodate}.

\subsection{Out-of-Distribution (OOD) Detection} \label{sec:ood}

Many recent approaches have focused on quantifying distributional uncertainty caused by a change in the input data distribution \cite{quinonero-candela_dataset_2009}. Approaches that are specifically focused on determining if an input falls outside of the input-data distribution are referred to as out-of-distribution (OOD) detection methods. These approaches are generally either (1)~classification-based \cite{liang_enhancing_2020,hsu_generalized_2020,liu_energy-based_2020}, (2)~density-based \cite{zong2018deep,kde,rezende_variational_2016,kingma_glow_2018,ren-2019}, (3)~distance-based \cite{lee-2018,Zaeemzadeh_2021_CVPR,Techapanurak_2020_ACCV,sun_out--distribution_2022}, or (4)~reconstruction-based \cite{xia-2015,gong_memorizing_2019,An2015VariationalAB,chen-2018,zenati_efficient_2019,Sabokrou,OCGAN}. While these methods do not fully capture the predictive uncertainty associated with DNNs, they can identify inputs that are drawn from a different distribution than those used to train the model. For these inputs, the model cannot be safely trusted to obtain the correct output. 

\subsection{Anomaly Detection \& Localization} \label{sec:anomaly}

A field of study that is closely related to OOD detection is anomaly detection and localization. The task of segmenting the particular pixels containing anomalies has become popular for identifying defects in industrial inspection \cite{tao_deep_2022}, flagging abnormalities in medical image analysis \cite{haber_anomaly_2022,narayanan_survey_2023}, and detecting abnormal behavior in surveillance applications \cite{anoopa_survey_2022}. Within the area of anomaly detection and localization, most approaches are (1)~reconstruction-based \cite{niethammer_unsupervised_2017,schlegl_f-anogan_2019,bergmann_improving_2019,zavrtanik_reconstruction_2021}, (2)~classification-based \cite{li_cutpaste_2021,zavrtanik_draem_2021}, or (3)~distance-based \cite{bergmann_uninformed_2020,del_bimbo_padim_2021,ishikawa_patch_2021,cohen_sub-image_2021}. While methods in uncertainty quantification, OOD detection, and anomaly localization can successfully capture different aspects of predictive uncertainty associated with DNNs, it is not clear how these methods should be used to develop safe and robust planning and control schemes.

\subsection{Uncertainty-Aware Navigation} \label{sec:uncertain_nav}

There are various approaches that seek to enable  short-term, local control under perception model predictive uncertainty. Existing approaches tend to rely on one of the UQ methods discussed in Section \ref{sec:uncertainty}, using either BNNs \cite{habli_improving_2021,arnez_quantifying_2022}, MC dropout \cite{nguyen_motion_2022}, or ensembles \cite{lotfi_uncertainty-aware_2023,triest_learning_2023,nguyen_uncertainty-aware_2023} to estimate uncertainty and inform the control scheme. As mentioned previously, while these methods can effectively estimate some aspects of predictive uncertainty, they are insufficient for quantifying uncertainty associated with OOD samples, which a UGV is bound to encounter when navigating previously unseen or dynamic environments. There has also been work that leverages uncertainty estimation approaches that relate more closely to the distance-based OOD detection methods \cite{tan_risk-aware_2021,virani_justification-based_2021} and classification-based anomaly detection methods \cite{seo_safe_2023,seo_scate_2023}. While these approaches do not fully capture the predictive uncertainty of the perception models, they are better equipped to handle inputs that differ significantly from those seen during training.

We develop a reconstruction-based method for predictive uncertainty estimation with the end goal of probabilistically safe control of a UGV. Our approach for uncertainty estimation has several benefits compared to existing methods. (1)~It captures multiple aspects of uncertainty and can effectively identify OOD or anomalous samples, unlike traditional UQ techniques.  (2)~In contrast to methods that simply seek to identify samples that differ significantly from the training set, our method is directly tied to the predictive uncertainty of the perception model used in a control task. Furthermore, rather than making a binary decision, as is typical in OOD and anomaly detection tasks, our method is probabilistic and can be incorporated into a probabilistically safe control scheme. (3)~Finally, our method can characterize uncertainty at a regional level, rather than considering uncertainty only for the image as a whole. This offers a greater amount of model explainability and provides more flexibility in designing an uncertainty-aware planning and control scheme. We show how our method for predictive uncertainty estimation can be used to evaluate the probability of error associated with vehicle trajectories and to select a preferred trajectory based on the desired level of conservativeness.\footnote{The code for reproducing our methods and results is available on GitHub: \url{https://github.com/sarapohland/parce-nav.git}.}

\section{PROBLEM FORMULATION}

In this work, we consider a UGV navigating a simulated lunar environment, in which robust visual navigation is very advantageous. During training, the perception model learns to distinguish between different regions in the environment (e.g., bumpy terrain, smooth terrain, inside crater, edge of crater, etc.) to navigate effectively. During evaluation, it encounters astronauts and human-made obstacles that were not seen during training. These unfamiliar obstacles are outlined in white in Figure \ref{fig:eval-scenarios} and are considered OOD because they were not present in the training set. The vehicle is expected to recognize regions in the environment that are unfamiliar to the perception model and continue to navigate effectively in the face of predictive uncertainty. To enable effective navigation, we develop a probabilistic competency estimation method for the perception model (Section \ref{sec:perception}), then use these competency estimates to develop competency-aware planning and control schemes (Section \ref{sec:navigation}). Additional details and parameters are provided in Appendices \ref{app:details} and \ref{app:params}.

\section{MODEL COMPETENCY ESTIMATION} \label{sec:perception}

We develop a probabilistic reconstruction-based method for competency estimation and evaluate the ability of this method to distinguish between correctly classified in-distribution (ID) samples, incorrectly classified ID samples, and out-of-distribution (OOD) samples (\S \ref{sec:overall-comp}). We then extend this approach to estimate model competency for regions in an image and evaluate the ability to distinguish between regions in ID samples, familiar regions in OOD samples, and unfamiliar regions in OOD samples (\S \ref{sec:regional-comp}). 

\subsection{Estimating Overall Model Competency} \label{sec:overall-comp}

Let $f$ be the true underlying model of the system from which our images are drawn and $\hat{f}$ be the predicted model (referred to as the perception model). For an input image, $\bm{X}$, the perception model aims to estimate the true class of the image, $f(\bm{X})$, from the set of all classes, $\mathcal{C}$. The competency of the model for this image is given by
\begin{equation}
    \rho(\bm{X}) := P\bigl(\{\hat{f}(\bm{X})=f(\bm{X})\}|\bm{X}\bigr).
\end{equation}

To simplify our notation, let $\hat{c}$ be the class predicted by the perception model (i.e., $\hat{f}(\bm{X})=\hat{c}$) such that
\begin{equation}
    \rho(\bm{X}) = P\bigl(\{f(\bm{X})=\hat{c}\}|\bm{X}\bigr).
\end{equation}

Often, the perception model uses the softmax function to obtain an estimate of the probability $P(\{f(\bm{X})=c)\}|\bm{X})$ for each class $c\in\mathcal{C}$. However, the perception model cannot truly estimate this probability because it is limited by the data contained in the training sample. It instead estimates the probability $P(\{f(\bm{X})=c)\}|\bm{X}, D)$, where $D$ is the event that the input image is in-distribution (i.e. drawn from the same distribution as the training samples). Let us then write the following lower bound on competency:
\begin{equation}
    \rho(\bm{X}) \geq 
P\bigl(D \cap \{f(\bm{X})=\hat{c})\}|\bm{X}\bigr).
\end{equation}

This lower bound can equivalently be expressed as
\begin{equation}
    \rho(\bm{X}) \geq P\bigl(\{f(\bm{X})=\hat{c}\}|\bm{X}, D\bigr)P(D|\bm{X}).
\end{equation}

We can assume that the perception model provides an estimate of the first probability. To estimate the second probability, we design an autoencoder to reconstruct the input image, training the autoencoder with the same images used to train the perception model. A holdout set is then used to estimate the distribution of the reconstruction loss for each class. It is assumed the reconstruction loss for a given class, $\mathcal{L}_c$, follows a Gaussian distribution with mean $\mu_c$ and standard deviation $\sigma_c$. Let $\ell(\bm{X})$ be the reconstruction loss for image $\bm{X}$. The probability this image is drawn from the same distribution as those in the training sample is given by
\begin{equation}
    P(D|\bm{X}) = \sum_{c\in\mathcal{C}} P(D|\{f(\bm{X})=c\})P(\{f(\bm{X})=c\}|\bm{X}).
\end{equation}

Assume now that the perception model provides an accurate estimate of $P(\{f(\bm{X})=c\}|\bm{X})$. We can estimate $P(D|\{f(\bm{X})=c\})$ as the probability that the reconstruction loss corresponding to image $\bm{X}$ aligns with the bottom $N\%$ of the training images. Because $\mathcal{L}_c$ is a Gaussian random variable, $N$ corresponds to a z-score, $z$, and we can estimate $P(D|\{f(\bm{X})=c\})$ as $\hat{p}_{D|C}$ in the following way:
\begin{align}
    \hat{p}_{D|C}
    &= P\bigl(\{\mathcal{L}_c > \ell(\bm{X}) - (\mu_c + z\sigma_c)\}\bigr) \\
    &= 1 - P\bigl(\{\mathcal{L}_c \leq \ell(\bm{X}) - (\mu_c + z\sigma_c)\}\bigr) \\
    &= 1 - F_{\mathcal{L}_c\sim\mathcal{N}(\mu_c,\sigma_c)}\bigl(\ell(\bm{X}) - \mu_c - z\sigma_c\bigr) \\
     &= 1 - F_{Z\sim\mathcal{N}(0,1)}\left(\frac{\ell(\bm{X}) - 2\mu_c - z\sigma_c}{\sigma_c}\right) \\
     &=: 1 - \phi\left(\frac{\ell(\bm{X}) - 2\mu_c}{\sigma_c} - z\right).
\end{align}

Let $\hat{p}_c$ be the probabilistic (softmax) output of the perception model corresponding to class $c\in\mathcal{C}$. We now have the following estimate of model competency:
\begin{equation}
    \hat{\rho}(\bm{X}) := \hat{p}_{\hat{c}} \sum_{c\in\mathcal{C}} \hat{p}_c\left(1 - \phi\left(\frac{\ell(\bm{X}) - 2\mu_c}{\sigma_c} - z\right)\right).
\end{equation}

We refer to $\hat{\rho}(\bm{X})$ as the \textit{overall competency score} for image $\bm{X}$. If the model is 100\% competent on the image, the probability that its prediction is correct is one. If it is entirely incompetent, the probability its prediction is correct is zero. The competency score is thus between zero and one, with higher scores corresponding to higher levels of competency\footnote{Additional details about this competency score and how it is calibrated are available on arXiv: \url{https://arxiv.org/abs/2411.16715}.}. 

\subsection{Analyzing Overall Model Competency Scores}

\begin{figure}[h]
    \centering
    \captionsetup{width=\columnwidth}
    \includegraphics[width=\columnwidth]{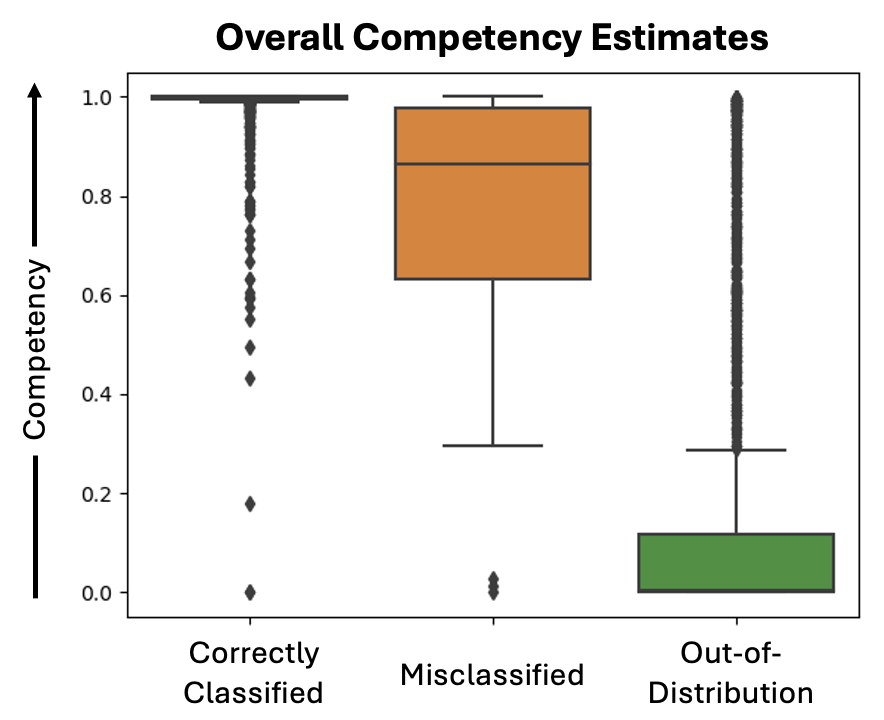}
    \caption{Distribution of overall model competency estimates for correctly classified ID samples (left), misclassified ID samples (middle), and OOD samples (right).}
    \label{fig:comp-overall}
\end{figure}

The distribution of competency scores for correctly classified ID images, misclassified ID images, and OOD images is shown in Figure \ref{fig:comp-overall}. Notice that the competency score is very close to one for nearly all correctly classified samples. It is below 0.3 for the majority of OOD images and is between 0.3 and 1.0 for nearly all misclassified ID samples. 

We compare our competency estimator to twelve existing methods for UQ and OOD detection. The capability of our method to distinguish between correctly classified and misclassified samples is comparable to existing methods, but our scoring method is much better at distinguishing between ID and OOD samples. Comparing our method to competing ones, the Kolmogorov–Smirnov (KS) distance between ID and OOD distributions is two times greater or more. When seeking to identify OOD samples from a set of ID samples, our method achieves an area under the receiver operating characteristic curve (AUROC) score of nearly one, and the false positive rate (FPR) is close to zero in both cases, which is an order of magnitude smaller than existing methods. In addition, we believe our scoring method is more intuitive than others, producing scores that are always between zero and one, as opposed to existing methods with no preset range. See Appendix \ref{app:overall} for a more thorough comparison of our competency score to existing methods.

\subsection{Estimating Regional Model Competency} \label{sec:regional-comp}

Suppose that, in addition to estimating the overall competency score for the image as a whole, we wish to estimate a regional competency score for each segmented region in the image. Now, instead of letting $\bm{X}$ be the entire input image, $\bm{X}$ is a segmented region of the image. In this work, segmented regions are determined by the Felzenszwalb segmentation algorithm \cite{felzenszwalb_efficient_2004}. We follow roughly the same procedures to estimate the probability that each region in the input image came from the same distribution as the training samples. In this approach, rather than designing an autoecoder to reconstruct the input image, we design an image inpainting model to reconstruct a missing segment of the input image and measure the average reconstruction loss over the pixels corresponding to that image segment. 


\subsection{Analyzing Regional Model Competency Maps}

\begin{figure}[t!]
    \centering
    \captionsetup{width=\columnwidth}
    \includegraphics[width=\columnwidth]{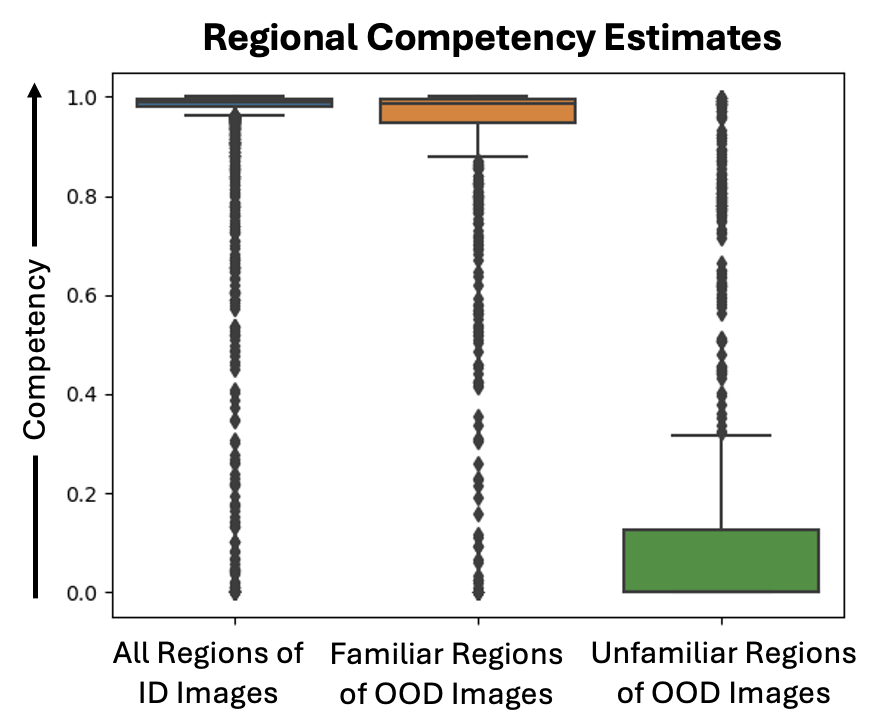}
    \caption{Distribution of regional competency estimates for regions in ID samples (left), familiar regions in OOD samples (middle), and unfamiliar regions in OOD samples (right).}
    \label{fig:comp-regional}
\end{figure}

\begin{figure}[h!]
    \centering
    \captionsetup{width=\columnwidth}
    \includegraphics[width=\columnwidth]{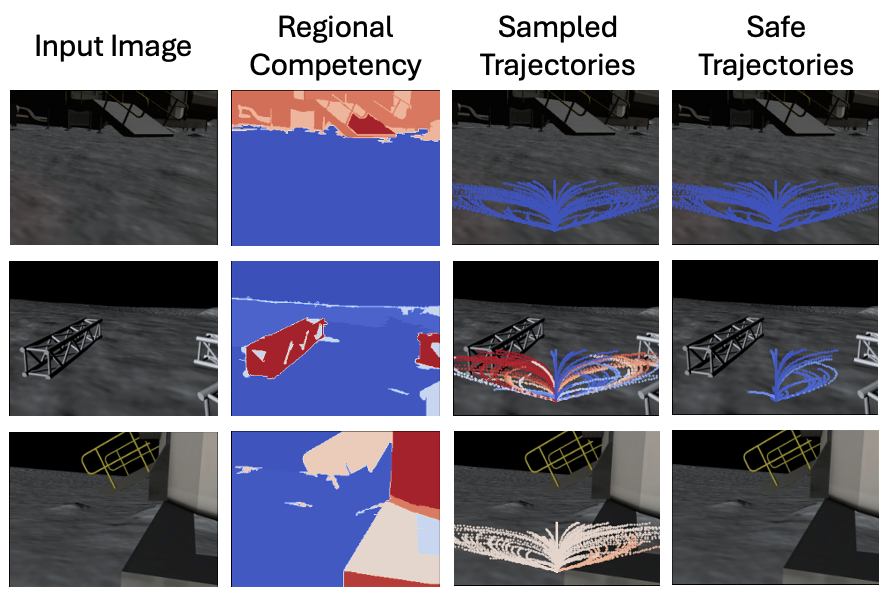}
    \caption{Three examples of competency-aware planning. \textit{Column 1:} Input images to perception model. \textit{Column 2:} Probabilistic regional competency estimates. Blue indicates high competency (around 1.0) and red indicates low competency (around 0.0). \textit{Column 3:} Sampled trajectories colored by associated minimum competency score. \textit{Column 4:} Trajectories deemed safe by the competency-aware planner.}
    \label{fig:planning}
\end{figure}

The distribution of competency scores for regions in ID images, familiar regions in OOD images, and unfamiliar regions in OOD images is shown in Figure \ref{fig:comp-regional}. Notice that the competency score is close to one for most of the familiar regions (in both ID and OOD images), while it is close to zero for most of the unfamiliar regions in OOD images. We can use these competency estimates to generate \textit{regional competency maps} (examples in column two of Figure \ref{fig:planning}).  

We compare our regional competency estimation method to seven existing methods for anomaly detection and localization. We find that our method outperforms existing methods when tasked with distinguishing between familiar and unfamiliar pixels, achieving the lowest FPRs, the highest KS distance between familiar and unfamiliar pixels in OOD images, the second highest KS distance between pixels in ID images and unfamiliar pixels in OOD images, and the second highest AUROC values. Although its performance is not significantly better than that of PaDiM \cite{del_bimbo_padim_2021}, PaDiM produces anomaly scores in a variable range, while our method generates probabilistic scores between 0 and 1, which are more interpretable and straightforward to incorporate into a planning framework. See Appendix \ref{app:regional} for a more thorough comparison of our regional model competency maps to existing methods.

\section{COMPETENCY-AWARE NAVIGATION} \label{sec:navigation}


We use the overall model competency estimates (\S \ref{sec:overall-comp}) and regional competency estimates (\S \ref{sec:regional-comp}) to design navigation schemes with varying levels of competency awareness (\S \ref{sec:planning}) and evaluate them across five scenarios (\S \ref{sec:eval}).





\subsection{Competency-Aware Path Planning} \label{sec:planning}

Given the vehicle’s current position and goal, along with an image of the environment from the vehicle’s perspective, the path planner determines the best trajectory to take from a set of sampled paths. To obtain a diverse set of dynamically feasible position paths, the planner samples action sequences within velocity control bounds, then predicts the vehicle's path under those actions using the estimated dynamics. The baseline planner, without any competency-awareness, selects the best path based on the progress it makes towards the goal and the vehicle orientation with respect to the goal position.

We develop five planners that utilize different competency information and strategies for responding to low competency. (1)~The planner that uses \textit{only overall} competency selects the best trajectory in the same way as the baseline planner when overall competency is high. If the competency score falls below some probability threshold, the planner resorts to a safe action sequence--backing up slightly then turning. (2)~The \textit{turning-based} planner that utilizes \textit{only regional} competency information checks if regions in the vicinity of the robot are associated with low perception model competency. If those regions fall below some probability threshold, the vehicle backs up slightly and turns away from low competency regions. (3)~The \textit{turning-based} planner that uses \textit{both overall and regional} competency checks both measures of competency before resorting to a safe response. (4)~The \textit{trajectory-based} planner that uses \textit{only regional} competency information maps sampled paths onto the regional competency map, determines the minimum competency value associated with each path, and removes those that fall below the threshold. The planner will then select the best path among those remaining. If there is no safe path, the vehicle will back up slightly and turn away from the low competency regions. (5)~The \textit{trajectory-based} planner that utilizes \textit{both overall and regional} competency behaves similarly but only considers regional competency when overall competency is low. Three examples of this trajectory-based planner with full competency-awareness are shown in Figure \ref{fig:planning}.



To follow the paths generated by the competency-aware path planner, we develop a reference-tracking linear-quadratic regulator (LQR) using a linearized model of vehicle dynamics. See Appendix \ref{app:details} for more details on our vehicle model, competency-aware path planner, and path-tracking controller.

\subsection{Evaluation of Navigation Performance} \label{sec:eval}

We evaluate the navigation capability of the full perception, planning, and control pipeline (Figure \ref{fig:overview}) with varying levels of competency awareness and different responses to low model competency. We compare (1)~a \textit{baseline} controller that uses no competency information, (2)~a \textit{turning-based} controller that uses \textit{only overall} competency scores, (3)~a \textit{turning-based} controller that uses \textit{only regional} competency maps, (4)~a \textit{trajectory-based} controller that uses \textit{only regional} competency information, (5)~a \textit{turning-based} controller that uses \textit{both} overall competency scores and regional competency maps, and (6)~a trajectory-based controller that uses \textit{both} overall and regional competency information. 

\begin{figure}[h!]
    \centering
    \captionsetup{width=\columnwidth}
    \includegraphics[width=\columnwidth]{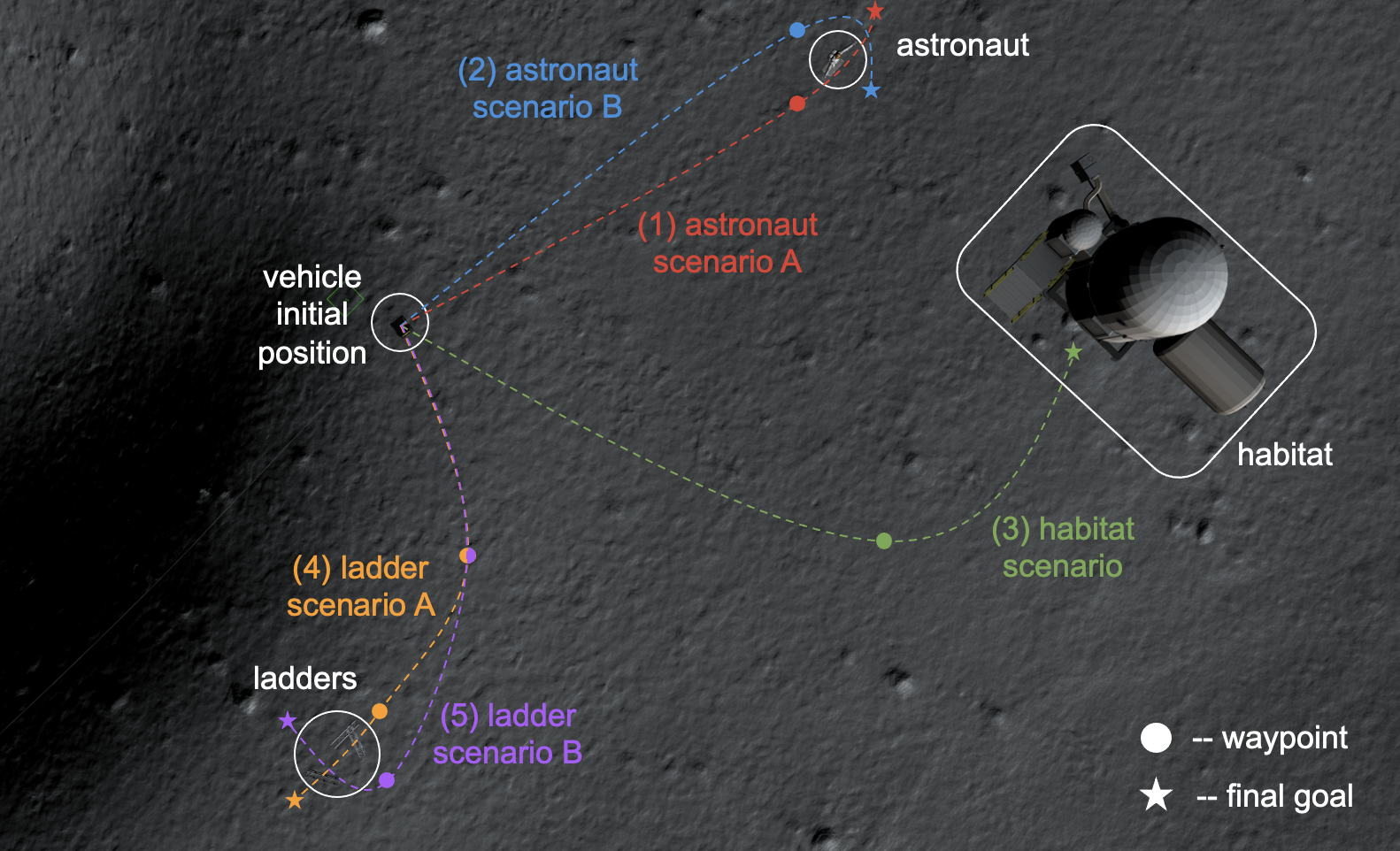}
    \caption{Five scenarios used to evaluate navigation performance: (1)~Drive from one side of astronaut to the other. (2)~Drive from back to front of astronaut. (3)~Drive up to front of habitat. (4)~Navigate around both sets of ladders. (5)~Navigate between two sets of ladders.}
    \label{fig:eval-scenarios}
\end{figure}

We conduct ten trials in each of the five navigation scenarios shown in Figure \ref{fig:eval-scenarios}. All of these scenarios are implemented in a simulated lunar environment built on Gazebo, and all evaluations are run in real time. We evaluate navigation performance based on: (1)~\textit{Success rate}: percentage of trials in which the vehicle reaches its goal within the allotted time. (2)~\textit{Timeout rate}: percentage of trials in which the vehicle fails to reach the goal in the allotted time. (3)~\textit{Collision rate}: percentage of trials in which the vehicle collides at least once. (4)~\textit{Navigation time}: average time required to navigate to the goal (in seconds). (5)~\textit{Path length}: average distance traveled by the vehicle to its goal (in meters). The results of these experiments are summarized in Table \ref{table:navigation}.

\begin{table}[h!]
\centering
\vspace{2mm}
\scalebox{0.75}{
\begin{tabular}{|cc|ccccc|}
\hline
Competency & Controller & Success  & Timeout  & Collision  & Navigation  & Path  \\
Awareness & Response &  Rate &  Rate &  Rate &  Time ($s$) &  Length ($m$)  \\
\hline
None & N/A & 62\% & 38\% & 70\%  & 69.61  & 35.34   \\
Overall & Turning & 10\% & 90\% & 4\%  & 82.86  & 38.20   \\
Regional & Turning & 90\% & 10\%  & 10\%  & 66.25  & 36.93   \\
Regional & Trajectory & 88\% & 12\%  & 8\%  & 66.61  & \textbf{34.32}   \\
Both & Turning & 88\%  & 12\%  & \textbf{2\%}  & 66.42  & 36.62  \\
Both & Trajectory & \textbf{100\%} & \textbf{0\%}  & 4\%  & \textbf{64.08}  & 34.74   \\
\hline
\end{tabular}
}
\caption{A comparison of (1)~a baseline controller with no competency awareness, (2)~a turning-based controller with only overall competency information, (3)~a turning-based controller with only regional competency information, (4)~a trajectory-based controller with only regional competency information, (5)~a turning-based controller with overall and regional competency information, and (6)~a trajectory-based controller with both types of competency information.
}
\label{table:navigation}
\end{table}

From Table \ref{table:navigation}, we notice that without competency-awareness, the collision rate for the baseline controller is very high. By adding overall competency-awareness, the collision rate is significantly reduced, but the navigation time increases notably, leading to a high timeout rate.
Adding regional competency information results in far higher rates of success, while maintaining low collision rates. Success rates are similar for all four controllers that use regional competency maps but are highest for the trajectory-based controller that uses both overall and regional competency information for planning. These results demonstrate the utility of perception model competency information for improving navigation performance around obstacles that are unfamiliar to the perception model. Furthermore, regional competency maps can significantly improve navigation efficiency. 

Again looking at Table \ref{table:navigation}, navigation times are similar across the four controllers that use regional competency maps, but the average time is slightly lower for the trajectory-based controller that uses both overall and regional information. Path lengths are also similar across the four controllers that use regional information, but they are lowest for the two trajectory-based approaches. These results demonstrate that a trajectory-based planning approach can further improve navigation efficiency in unfamiliar or changing environments. 

A more nuanced evaluation of navigation schemes across different scenarios is provided in Appendix \ref{app:nav}. To see a visual comparison of the baseline controller, the controller using only overall competency, the turning-based controller utilizing both overall and regional competency, and the trajectory-based controller utilizing both types of competency information, please view our demo video.\footnote{\label{video} A visual comparison of navigation performance across different planners is available on YouTube: \url{https://youtu.be/mUB2MDQZObU}.}

\section{CONCLUSIONS}

We aim to improve the safety of a UGV navigating according to a perception-based system in  a complex environment with obstacles that are unfamiliar to the perception model. We believe that integrating information about the predictive uncertainty of the perception model into the navigation scheme will improve the safety and performance of the system. With this idea, we propose a novel overall competency score that can distinguish between correctly classified ID samples, misclassified ID samples, and OOD samples. We expand on this approach to develop a regional competency map that can visualize regions in an image that are familiar to the perception model and those that are not. We then explore how this competency information can be integrated into a planning and control scheme for UGV navigation.

We find that, in general, competency information is useful for collision-free navigation around obstacles that are unfamiliar to the perception model. Regional competency information is particularly valuable for efficient navigation. Furthermore, we find that simply resorting to a safe action response (e.g., backing up slightly and turning) when the perception model is incompetent is often sufficient to enable collision-free navigation. However, a trajectory-based response (e.g., evaluating proposed vehicle trajectories based on regional model competency and avoiding paths through low competency regions) improves navigation efficiency. Overall, we demonstrate the utility in integrating our competency estimation method into the navigation framework.

\section{LIMITATIONS \& FUTURE WORK}

While the integration of competency-awareness into the navigation framework greatly reduces the number of collisions with unfamiliar obstacles, collisions do still occur for a couple reasons. The field-of-view of the vehicle's front-facing camera is quite limited, and the vehicle currently has no visual memory, so it will collide with obstacles that it cannot see. It would be useful for the vehicle to maintain some memory of competency estimates for nearby regions that are no longer visible. The competency estimators are also imperfect, and errors in competency estimates carry through the perception, planning, and control pipeline, which can result in collisions. It would be interesting to explore how these errors propagate through the navigation pipeline. As a final note, we are currently employing a relatively simple graph-based image segmentation algorithm, whose limitations can affect the performance of our regional competency estimator. Future work could explore domain-specific segmentation methods or emerging foundation models for image segmentation to enhance performance.

There are also various other avenues for future work. First, there is currently an implicit assumption that spatial features of an image, corresponding to unfamiliar obstacles, cause the reduction in perception model competency. It would be useful to develop a more general framework for understanding model competency in cases where predictive uncertainty is not due to some particular region in the image/environment. Second, the competency estimates are currently formulated for classification problems. One could generalize these formulations to be used for other tasks. In addition, this work focused on integrating competency-awareness into a local planning and control framework. It would be interesting to develop a global planner with competency-awareness. Additionally, the current response to perception model competency is to avoid regions associated with low competency. To enable continual learning within a dynamic or unfamiliar environment, it would be useful to intentionally and cautiously explore unfamiliar regions, rather than simply avoid them. It would be interesting to consider methods for safe exploration and decision frameworks to trade off between avoidance and exploration. As a final note, while robust visual navigation is critical in lunar environments, robustness to perception uncertainty is important in other domains as well. One might consider the application of this work to other areas of field robotics.

\bibliographystyle{IEEEtran}
\bibliography{citations}

\clearpage
\newpage
\appendix

\subsection{ADDITIONAL IMPLEMENTATION DETAILS} \label{app:details}

In this section, we provide some additional details on the implementation of our competency-aware navigation scheme.

\textbf{\textit{Model of Vehicle Dynamics:}} We model our vehicle as a non-linear discrete time system whose state, $\vec{x}\in\mathbb{R}^5$, can be described in terms of its X-position ($x$), Y-position ($y$), heading ($\theta$), linear velocity ($v$), and turn rate ($\omega$). The input, $\vec{u}\in\mathbb{R}^2$ to the system is the desired linear velocity or throttle ($t$) and desired turn rate or steering command ($s$). The state and input are expressed as 
\[
\vec{x}=
\begin{bmatrix}
    x & y & \theta & v & \omega
\end{bmatrix}^T
\hspace{2mm}\text{and}\hspace{2mm}
\vec{u}=
\begin{bmatrix}
    t & s
\end{bmatrix}^T
\hspace{2mm}\text{respectively.}
\]

The vehicle is then described by a discrete time-varying model that is very similar to the Dubin's car model:
\[
\vec{x}_{k+1} = \bm{A_k}\vec{x}_k + \bm{B}\vec{u}_k,\ 
\text{where}
\]
\[
\bm{A_k} = 
\begin{bmatrix}
    1 & 0 & 0 & \Delta t\cos{\Tilde{\theta}_k} & 0 \\
    0 & 1 & 0 & \Delta t\sin{\Tilde{\theta}_k} & 0 \\
    0 & 0 & 1 & 0 & \Delta t \\
    0 & 0 & 0 & 1-\alpha & 0 \\
    0 & 0 & 0 & 0 & 1-\beta
\end{bmatrix}
\hspace{5mm}
\bm{B} = 
\begin{bmatrix}
    0 & 0 \\
    0 & 0 \\
    0 & 0 \\
    \alpha & 0 \\
    0 & \beta \\
\end{bmatrix}.
\]

In the above equation, $\Delta t$ is the time step, and the parameters $\alpha$ and $\beta$ are estimated from data. To generate a linear approximation for the vehicle dynamics, we use an estimate of the vehicle's heading, $\Tilde{\theta}_k$, at future time steps.

\vspace{2mm}

\textbf{\textit{Competency-Aware Path Planning:}} The path planner samples $N$ action sequences within the velocity control bounds $[\vec{u}_{min}, \vec{u}_{max}]$ over a time horizon of $H$, then estimates the state of the vehicle at future time steps under each action sequence (based on the dynamics model from Section \ref{sec:dynamics}). It then removes paths that exit the field-of-view of the vehicle. The baseline planner, without any competency-awareness, selects the best path among the sampled trajectories based on the progress it makes towards the goal and the vehicle orientation with respect to the goal. Let $\vec{x}_{0:H}$ be the state of the vehicle across the entire planning horizon and $\vec{x}_{goal}$ be the $(x, y)$ goal position. The quality of a sampled path can be summarized by the following cost function:
\begin{align*}
    & c(\vec{x}_{0:H}, \vec{x}_{goal}) = \\
    &\hspace{5mm} \alpha_{ori} \left(\tan^{-1}\left(\frac{\vec{x}_{goal}[1]-\vec{x}_H[1]}{\vec{x}_{goal}[0]-\vec{x}_H[0]}\right) - \vec{x}_H[2]\right)^2 \\
    &\hspace{10mm}  + \alpha_{goal\_x} |\vec{x}_{goal}[0]-\vec{x}_H[0]| \\
    &\hspace{10mm}  + \alpha_{goal\_y} |\vec{x}_{goal}[1]-\vec{x}_H[1]|.
\end{align*}

The best path among those sampled minimizes this cost function. The weighting parameters ($\alpha_{ori}$, $\alpha_{goal_x}$, and $\alpha_{goal_y}$) are tuned to get good planning performance, and the chosen values are provided in Appendix \ref{app:params}.

For the trajectory-based path planners that utilize regional competency information, we project the estimated position paths onto the current image representing the vehicle's environment. We evaluate the minimum competency corresponding to the space occupied by the vehicle at each time step along the trajectory and remove paths whose minimum competency is below the competency threshold. We then select the sampled path with the minimum cost among the remaining paths. If there are no safe trajectories, the vehicle resorts to a safe action response.

\vspace{2mm}

\textbf{\textit{Path-Tracking Controller:}} To follow the paths generated by the competency-aware path planner, we develop a reference-tracking linear-quadratic regulator (LQR). We seek the control inputs $\vec{u}_0,\vec{u}_1,\hdots,\vec{u}_{N-1}$ that minimize
\small
\[
\sum_{k=0}^{N-1} (\vec{x}_k-\vec{x}_k^{ref})^T\bm{Q}(\vec{x}_k-\vec{x}_k^{ref}) + (\vec{u}_k-\vec{u}_k^{ref})^T\bm{R}(\vec{u}_k-\vec{u}_k^{ref})
\]
\normalsize
constrained by the initial state of the vehicle and the linearized dynamics. The reference state, $\vec{x}_k^{ref}$, and reference input, $\vec{u}_k^{ref}$, are obtained from the competency-aware path planner (Section \ref{sec:planning}). The parameters of the state deviation cost matrix, $\bm{Q}$, and input deviation cost matrix, $\bm{R}$, are tuned to get good performance, and the chosen values are provided in Appendix \ref{app:params}. Solving this optimization problem, we find the optimal input at time step $k$ to be
\[
\vec{u}^*_k = -\bm{K_k}(\vec{x}_k-\vec{x}_k^{ref}) + \vec{u}_k^{ref}, \text{ where }
\]
\[
\bm{K_k} = (\bm{R}+\bm{B}^T\bm{P_{k+1}}\bm{B})^{-1}\bm{B}^T\bm{P_{k+1}}\bm{A_k}
\]
\[
\bm{P_k} = \bm{Q} + \bm{K_k}^T\bm{R}\bm{K_k} + (\bm{A_k}-\bm{B}\bm{K_k})^T\bm{P_{k+1}}(\bm{A_k}-\bm{B}\bm{K_k}).
\]

Note that $\bm{P_N}$ is simply given by the state cost matrix, $\bm{Q}$.

\subsection{PARAMETERS FOR EVALUATION} \label{app:params}

In Table \ref{table:params}, we provide some of the parameter values chosen to evaluate our competency estimation method (Section \ref{sec:perception}) and competency-aware navigation scheme (Section \ref{sec:navigation}). Additional information about a number of these parameters is provided in our configurations \href{https://github.com/sarapohland/parce-nav/blob/main/navigation/control/configs/README.md}{README}.

\begin{table}[h]
\renewcommand{\arraystretch}{1.5}
\centering
\vspace{2mm}
\begin{tabular}{|c|c|c|}
\hline
\renewcommand{\arraystretch}{2} 
Type & Parameter & Value \\
\hline
\multirow{4}{*}{\shortstack{Competency \\ Estimation}} 
& Overall Confidence Value ($N_{ovl}$) & 95 \\
& Regional Confidence Value ($N_{reg}$) & 95 \\
& Overall Competency Threshold ($\eta_{ovl}$) & $0.8$ \\
& Regional Competency Threshold ($\eta_{reg}$) & $0.8$ \\
\hline
\multirow{3}{*}{\shortstack{Dynamics \\ Model}} 
& Time Step ($\Delta t$) & $0.1\ s$ \\
& Linear Velocity Factor ($\alpha$) & 0.26 \\
& Turn Rate Factor ($\beta$) & 0.35 \\
\hline
\multirow{6}{*}{\shortstack{Path \\ Sampling}} 
& Linear Velocity Lower Bound ($\vec{u}_{min}[0]$) & $0.0\ m/s$ \\
& Linear Velocity Upper Bound ($\vec{u}_{max}[0]$) & $0.8\ m/s$ \\
& Turn Rate Lower Bound ($\vec{u}_{min}[1]$) & $-0.4\ rad/s$ \\
& Turn Rate Upper Bound ($\vec{u}_{max}[1]$) & $0.4\ rad/s$ \\
& Number of Sampled Paths ($N$) & $128$ \\
& Time Horizon of Sampled Paths ($H$) & $60$ \\
\hline
\multirow{3}{*}{\shortstack{Path \\ Reward}} 
& Angle from Goal Cost ($\alpha_{ori}$) & $3.0$ \\
& Distance from $X$ Goal Cost ($\alpha_{goal\_x}$) & $1.0$ \\
& Distance from $Y$ Goal Cost ($\alpha_{goal\_y}$) & $1.5$ \\
\hline
\multirow{7}{*}{\shortstack{Path \\ Tracking}} 
& $X$ Position Deviation Cost ($\bm{Q}[0,0]$) & $1.0$ \\
& $Y$ Position Deviation Cost ($\bm{Q}[1,1]$) & $1.0$ \\
& Heading Deviation Cost ($\bm{Q}[2,2]$) & $2.0$ \\
& Linear Velocity Deviation Cost ($\bm{Q}[3,3]$) & $0.5$ \\
& Turn Rate Deviation Cost ($\bm{Q}[4,4]$) & $0.0$ \\
& Linear Velocity Input Cost ($\bm{R}[0,0]$) & $0.1$ \\
& Turn Rate Input Cost ($\bm{R}[1,1]$) & $0.1$ \\
\hline
\multirow{2}{*}{\shortstack{Safety \\ Response}} 
& Vehicle Backup Time & $1.0\ s$ \\
& Vehicle Turn Time & $1.0\ s$ \\
\hline
\end{tabular}
\caption{Parameter values chosen for our experiments.
}
\label{table:params}
\end{table}

\subsection{ANALYSIS OF OVERALL COMPETENCY SCORE} \label{app:overall}

In this section, we provide additional analysis of our overall model competency estimation method (Section \ref{sec:overall-comp}), in comparison to existing methods for uncertainty quantification (Section \ref{sec:uncertainty}) and OOD detection (Section \ref{sec:ood}).

\begin{table*}[h!]
\centering
\renewcommand{\arraystretch}{1.2} 
\begin{tabular}{|c|r|rrr|rrr|rrr|}
\hline
\multirow{2}{*}{Method} & \multirow{2}{*}{\shortstack{Computation \\ Time (sec) $\downarrow$}} & \multicolumn{3}{c|}{Correct vs. Incorrect} & \multicolumn{3}{c|}{Correct vs. OOD} & \multicolumn{3}{c|}{Incorrect vs. OOD} \\
\cline{3-5} \cline{6-8} \cline{9-11}
 & & Dist. $\uparrow$ & AUROC $\uparrow$ & FPR $\downarrow$ & Dist. $\uparrow$ & AUROC $\uparrow$ & FPR $\downarrow$ & Dist. $\uparrow$ & AUROC $\uparrow$ & FPR $\downarrow$ \\
\hline
\renewcommand{\arraystretch}{1} 
Softmax     &             \textbf{0.0001} &                  0.72 &        0.91 &      1.00 &       0.45 &        0.76 &      1.00 &       0.13 &        0.76 &      1.00 \\
Temperature \cite{guo} &             0.0002 &                  0.72 &        0.91 &      1.00 &       0.45 &        0.76 &      1.00 &       0.13 &        0.75 &      1.00 \\
Entropy     &             0.0002 &                  0.71 &        0.88 &      1.00 &       0.45 &        0.79 &      1.00 &       0.19 &        0.78 &      1.00 \\
Energy \cite{liu_energy-based_2020}      &             0.0002 &                  0.71 &        0.90 &      1.00 &       0.45 &        0.75 &      1.00 &       0.19 &        0.75 &      1.00 \\
MC Dropout \cite{dropout}     &             0.2950 &                  0.73 &        0.92 &      0.49 &       0.46 &        0.81 &      0.64 &       0.02 &        0.80 &      0.65 \\
Ensemble \cite{lakshminarayanan_simple_2017}    &             0.0408 &                  \textbf{0.80} &        \textbf{0.93} &      \textbf{0.28} &       0.49 &        0.79 &      1.00 &       0.01 &        0.78 &      1.00 \\
ODIN \cite{liang_enhancing_2020}        &             0.0955 &                  0.02 &        0.40 &      1.00 &       0.03 &        0.48 &      1.00 &       0.19 &        0.49 &      1.00 \\
KL-Matching \cite{kl_matching} &             0.0005 &                  0.65 &        0.80 &      0.83 &       0.44 &        0.74 &      0.88 &       0.22 &        0.74 &      0.88 \\
OpenMax \cite{openmax}     &             0.0017 &                  0.61 &        0.87 &      0.50 &       0.31 &        0.72 &      0.81 &       0.11 &        0.71 &      0.81 \\
Mahalanobis \cite{lee-2018} &             0.0722 &                  0.53 &        0.80 &      0.58 &       0.45 &        0.81 &      0.58 &       0.20 &        0.80 &      0.58 \\
k-NN \cite{sun_out--distribution_2022}         &             0.0280 &                  0.59 &        0.83 &      0.41 &       0.61 &        0.87 &      0.37 &       0.29 &        0.87 &      0.38 \\
DICE \cite{dice}        &             0.0101 &                  0.61 &        0.87 &      0.50 &       0.31 &        0.72 &      0.81 &       0.11 &        0.71 &      0.81 \\
\hline
PaRCE (Ours)       &             0.0300 &                  0.68 &        0.88 &      0.59 &       \textbf{0.89} &        \textbf{0.99} &      \textbf{0.08} &       \textbf{0.70} &        \textbf{0.99} &      \textbf{0.08} \\
\hline
\end{tabular}
\caption{A comparison of various scoring methods to measure overall model competency. Methods are evaluated based on their computation time and ability to distinguish between correctly classified, incorrectly classified, and OOD samples.}
\label{tab:overall}
\end{table*}

\textbf{\textit{Baselines:}} We compare our overall model competency estimation method against various existing methods for quantifying uncertainty and detecting OOD inputs. In particular, we compare our approach to the Maximum Softmax Probability (MSP) baseline, the calibrated MSP with Temperature Scaling \cite{guo}, the Entropy of the softmax probabilities, Monte Carlo (MC) Dropout \cite{dropout}, Ensembling \cite{lakshminarayanan_simple_2017}, ODIN \cite{liang_enhancing_2020}, the Energy Score \cite{liu_energy-based_2020}, KL-Matching \cite{kl_matching}, OpenMax \cite{openmax}, the Mahalanobis Distance \cite{lee-2018}, k-Nearest Neighbors (k-NN) \cite{sun_out--distribution_2022}, and DICE \cite{dice}. These methods serve as a representative selection of the many existing approaches and are often used as baselines in other works. A number of these methods were implemented with the help of the PyTorch-OOD library \cite{kirchheim2022pytorch}. Note that we do not compare to BNN methods because we do not place restrictions on the underlying model architecture used for classification. We also did not compare to existing reconstruction-based OOD detection methods because we found these approaches to be quite memory-intensive.

\textbf{\textit{Metrics:}} We evaluate these scoring methods based on their computation time, their ability to distinguish between correctly classified and misclassified samples, their ability to distinguish between correctly classified and OOD samples, and their ability to distinguish between misclassified and OOD samples. To evaluate the ability to distinguish between sets of samples using each scoring method, we consider the distance between score distributions using the Kolmogorov–Smirnov (KS) test, the extent of overlap between distributions as measured by the area under the receiver operating characteristic curve (AUROC), and the detection error determined by the False Positive Rate (FPR) at a 95\% True Positive Rate (TPR), where a true positive indicates the correct identification of a misclassified or OOD sample.

\begin{figure*}[h!]
    \centering
    \captionsetup{width=\textwidth}
    \includegraphics[width=\textwidth, height=10cm]{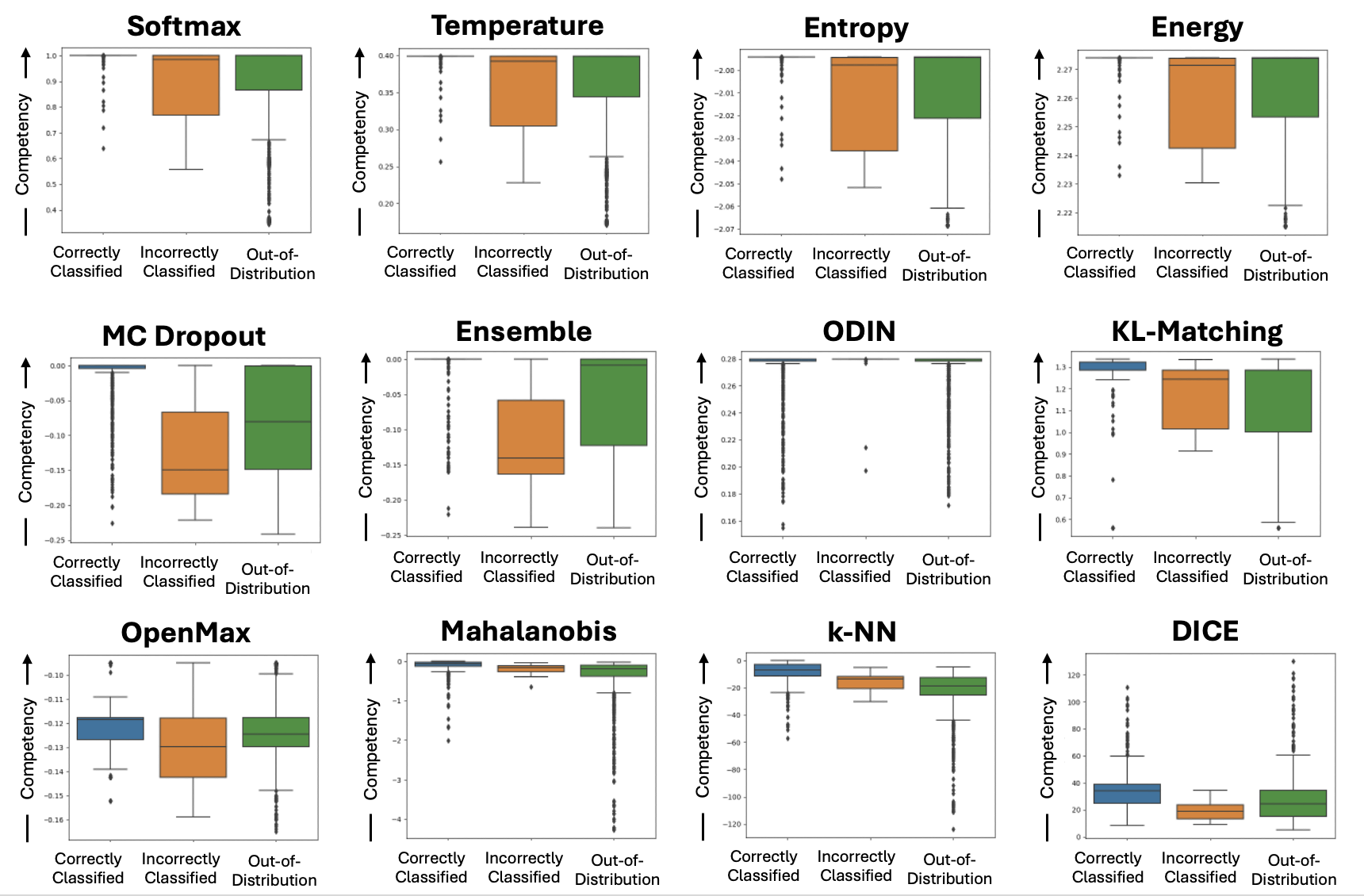}
    \caption{A comparison of the score distributions for correctly classified, incorrectly classified, and OOD samples under various existing uncertainty quantification and OOD detection methods. The results for our method are displayed in Figure \ref{fig:comp-overall}.}
    \label{fig:overall-compare}
\end{figure*}

\textbf{\textit{Results:}} The results are summarized in Table \ref{tab:overall}. We also display the distribution of competency scores for correctly classified, misclassified, and OOD samples in Figure \ref{fig:overall-compare}.

\textbf{\textit{Analysis:}} Notice from Table \ref{tab:overall} that, as expected, the Maximum Softmax method boasts the fastest computation time, and other methods that perform another simple function on the probability outputs of the perception model--Temperature Scaling, Entropy, and Energy--are also very fast. Looking at Figure \ref{fig:overall-compare}, we notice that for all four of these simple scoring functions, the correctly classified samples are almost always assigned a very high score, indicating high confidence in the label of these samples. In many cases, misclassified and OOD samples are assigned lower scores, so there is there is a reasonable separation between the score distributions of correctly classified and misclassified samples, as well as those of correctly classified and OOD samples (as evidenced by the KS Distances and AUROCs in Table \ref{tab:overall}). However, it is also very common for misclassified and OOD samples to be assigned equally high scores. Because of this, the FPR at a 95\% TPR is 1.0 across all cases, indicating that in order to detect 95\% of misclassified and OOD samples, we would incorrectly detect all correctly classified samples as well.

From Table \ref{tab:overall}, we can also see that Ensembling best distinguishes between correctly classified and misclassified samples across all metrics. However, it does not do as well at distinguishing between in-distribution (ID) samples (correctly or incorrectly classified) and OOD samples. Notice also from Figure \ref{fig:overall-compare} that the scores for OOD samples are often higher than those for misclassified ID samples. We see very similar results for MC dropout, which aligns with others' findings that traditional uncertainty quantification methods (dicussed in Section \ref{sec:uncertainty}) cannot be expected to appropriately assign confidence scores for OOD inputs \cite{schwaiger_is_nodate}. It is also worthwhile to note that MC dropout is significantly slower than all other methods. The evaluation time for Ensembling is not particularly high, but this method requires training multiple models, which is quite time consuming.

Looking again at Table \ref{tab:overall} and Figure \ref{fig:overall-compare}, aside from ODIN, which does quite poorly on this particular dataset, the remaining methods perform comparably when distinguishing between correctly classified and misclassified samples. In terms of distinguishing between ID samples (correctly or incorrectly classified) and OOD samples, the k-Nearest Neighbors method outperforms all existing methods.

Comparing the distribution of scores for our method (Figure \ref{fig:comp-overall}) to those of the existing methods (Figure \ref{fig:overall-compare}), our competency score provides a clear distinction between correctly classified, misclassified, and OOD samples. While the capability of our method to distinguish between correctly classified and misclassified samples is comparable to existing methods, our scoring method is much better at distinguishing between ID and OOD samples. From Table \ref{tab:overall}, we see that the KS distance between the distribution of correctly classified samples and that of OOD samples is nearly twice as high for our method compared to most existing methods we considered. When comparing the distribution of misclassified samples to that of OOD samples, this difference is far greater, with our method achieving distances no less than 2.4 times greater. When seeking to identify OOD samples from a set of correctly classified or misclassified samples, our method achieves an AUROC score of nearly one, and the FPR for the threshold that achieves a 95\% TPR is close to zero in both cases. These AUROC scores are much greater than those obtained by existing methods, and the FPRs are an order of magnitude smaller than most existing methods we considered. This indicates that our method is effective at estimating perception model competency for OOD inputs.

Finally, it is worthwhile to note that our scoring method is more intuitive than other methods. With our method, the competency score is always between zero and one, with higher scores corresponding to higher levels of competency. We find that for this dataset, the competency score is very close to one for nearly all correctly classified samples, it is below 0.3 for the majority of OOD images, and it is between 0.3 and 1.0 for nearly all misclassified ID samples (Figure \ref{fig:comp-overall}). With the exception of the Maximum Softmax score, which also has a predefined range of zero to one, none of the other methods limit scores to a set range (although a number are lower- or upper-bounded by zero). This makes it more challenging to identify an appropriate threshold for detecting OOD data without having this data available during training. 

As a final note, the average computation time for our method is comparable to other high-performing methods, obtaining a competency score for a given input within 0.03 seconds on average. This is fast enough to be useful in most decision-making pipelines, such as those used for autonomous navigation in unstructured environments.

\subsection{ANALYSIS OF REGIONAL COMPETENCY MAP} \label{app:regional}

\begin{table*}[h!]
\centering
\renewcommand{\arraystretch}{1.2} 
\begin{tabular}{|c|r|rrr|rrr|rrr|}
\hline
\multirow{2}{*}{Method} & \multirow{2}{*}{\shortstack{Computation \\ Time (sec) $\downarrow$}} & \multicolumn{3}{c|}{ID All vs. OOD Unfamiliar} & \multicolumn{3}{c|}{OOD Familiar vs. OOD Unfamiliar} \\
\cline{3-5} \cline{6-8}
 & & Dist. $\uparrow$ & AUROC $\uparrow$ & FPR $\downarrow$ & Dist. $\uparrow$ & AUROC $\uparrow$ & FPR $\downarrow$ \\
\hline
\renewcommand{\arraystretch}{1} 
DRAEM \cite{zavrtanik_draem_2021} &               1.315 &      0.293 &       0.633 &     0.989 &      0.367 &       0.641 &     0.989 \\
FastFlow \cite{fastflow} &               0.084 &      0.844 &       0.971 &     0.142 &      0.768 &       0.963 &     0.154 \\
PaDiM \cite{del_bimbo_padim_2021}        &               0.053 &      0.899 &       \textbf{0.984} &     0.056 &      0.811 &       \textbf{0.977} &     0.078 \\
PatchCore \cite{patchcore} & 1.656 &      \textbf{0.936} &       0.592 &     0.817 &      0.777 &       0.576 &     0.848 \\
Reverse Distillation \cite{reverse} &               0.349 &      0.790 &       0.915 &     0.160 &      0.698 &       0.906 &     0.182 \\
Regional KDE \cite{rkde} &               2.309 &      0.404 &       0.654 &     1.000 &      0.149 &       0.631 &     1.000 \\
Student-Teacher \cite{stfpm} &               \textbf{0.049} &      0.797 &       0.958 &     0.281 &      0.750 &       0.955 &     0.277 \\
\hline
PaRCE (Ours) &               0.418 &      0.901 &       0.976 &     \textbf{0.053} &      \textbf{0.875} &       0.976 &     \textbf{0.058} \\
\hline
\end{tabular}
\caption{A comparison of various mapping methods to measure and visualize regional model competency. Methods are evaluated based on their computation time and ability to distinguish between unfamiliar and familiar pixels in an image.}
\label{tab:regional}
\end{table*}

In this section, we provide additional analysis of our regional model competency estimation method (Section \ref{sec:regional-comp}), in comparison to existing methods for anomaly detection and localization that generate anomaly maps (Section \ref{sec:anomaly}).

\textbf{\textit{Baselines:}} We compare our regional competency estimator to various existing methods for anomaly detection and localization. In particular, we compare our approach to DRAEM \cite{zavrtanik_draem_2021}, FastFlow \cite{fastflow}, PaDiM \cite{del_bimbo_padim_2021}, PatchCore \cite{patchcore}, Reverse Distillation \cite{reverse}, Region-Based Kernel Density Estimation (KDE) \cite{rkde}, and Student-Teacher Feature Pyramid Matching \cite{stfpm}. These are several of the state-of-the-art anomaly detection algorithms that generate anomaly maps similar to our regional competency maps. All methods were implemented with the help of the Anomalib library \cite{akcay2022anomalib}.

\textbf{\textit{Metrics:}} We evaluate these anomaly maps based on their computation time, their ability to distinguish between regions in in-distribution (ID) images and unfamiliar regions in OOD images, and their ability to distinguish between familiar regions in OOD images and unfamiliar regions in OOD images. Familiar regions are all of the pixels that occupy image structures that exist in the training set, and unfamiliar pixels are those that occupy structures that were not present during training. To evaluate the ability to distinguish between sets of samples using each mapping method, we consider the distance between score distributions using the KS test, the extent of overlap between distributions as measured by the AUROC, and the detection error determined by the FPR at a 95\% TPR, where a true positive indicates the correct identification of a pixel in an unfamiliar region.

\begin{figure}[ht]
    \begin{minipage}{\textwidth}
        \centering
        \captionsetup{width=\textwidth}
        \includegraphics[width=\textwidth, height=7cm]{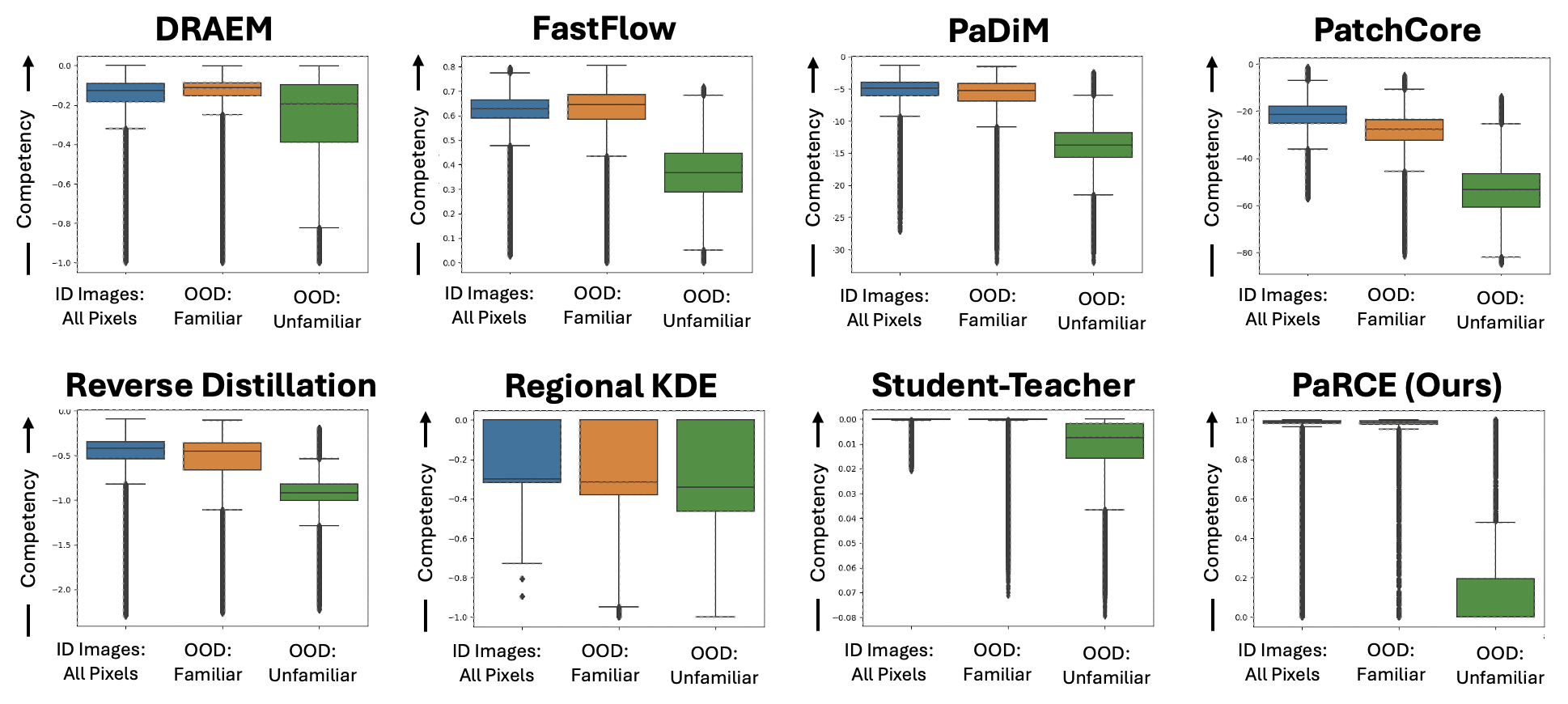}
        \caption{A comparison of the score distributions for all pixels in in-distribution images, familiar pixels in OOD images, and unfamiliar pixels in OOD images across various anomaly localization methods and our regional competency approach.}
        \label{fig:regional-compare}
     \end{minipage}
     
     \vspace{15mm}
     
     \begin{minipage}{\textwidth}
        \centering
        \captionsetup{width=\textwidth}
        \includegraphics[width=\textwidth, height=7cm]{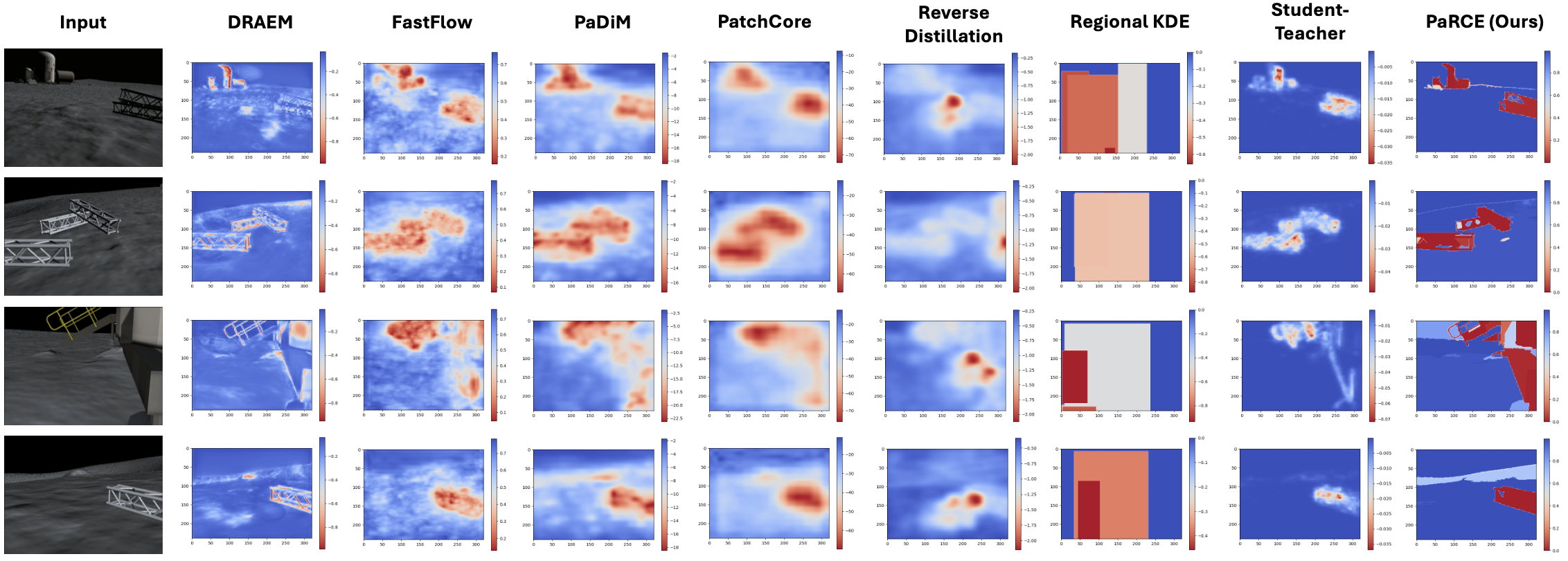}
        \caption{A comparison of predicted competency maps across various anomaly localization methods and our regional approach.}
        \label{fig:regional-maps}
     \end{minipage}
\end{figure}

\textbf{\textit{Results:}} The results are summarized in Table \ref{tab:regional}. We also display the distribution of competency scores for pixels in ID images, familiar pixels in OOD images, and unfamiliar pixels in OOD images in Figure \ref{fig:regional-compare}. In addition, Figure \ref{fig:regional-maps} displays the competency maps generated by each method for four example inputs in our lunar dataset. 

\textbf{\textit{Analysis:}} Comparing the eight methods for generating competency maps, there is quite a range in computation times (see Table \ref{tab:regional}). Student-Teacher Feature Pyramid Matching is the fastest method, but PaDiM and FastFlow run at comparable speeds. Reverse Distillation and our method run significantly slower but are still fast enough for many decision-making problems. DRAEM, PatchCore, and Regional KDE are much slower, requiring over a second to estimate a competency map. These methods would need to be more efficient to be useful in a planning and control pipeline.

While most methods we consider generate pixel-wise maps, Regional KDE generates bounding boxes with associated scores (see Figure \ref{fig:regional-maps} for examples). This provides much less granularity in distinguishing between familiar and unfamiliar regions, and there is little difference in the distribution of scores for this method (see Figure \ref{fig:regional-compare}). This results in low KS distances between familiar and unfamiliar pixels, as well as very high FPRs (see Table \ref{tab:regional}).

While DRAEM generates pixel-wise maps that seem to somewhat correlate with anomalous regions (see Figure \ref{fig:regional-maps}) and scores associated with unfamiliar regions are lower than those of familiar regions on average (see Figure \ref{fig:regional-compare}), this method does not perform very well on this dataset. There is not a huge difference in the distribution of scores, resulting in low KS distances, low AUROC values, and very high FPRs.

From the examples provided in Figure \ref{fig:regional-maps}, PatchCore appears to generate reasonable competency maps, and there is some difference between the scores assigned to familiar versus unfamiliar pixels. The KS distance between pixels in in-distribution images and unfamiliar pixels in OOD images is highest for this method, and the KS distance between familiar and unfamiliar pixels in OOD images is reasonably high as well. However, the AUROC values are lowest for this method, and the FPR at a 95\% TPR is very high.

From the examples in Figure \ref{fig:regional-maps} and boxplot in Figure \ref{fig:regional-compare}, we see the Student-Teacher Feature Pyramid Matching method tends to assign the maximum competency score to familiar pixels with very little variation, while unfamiliar pixels are often assigned lower scores. This results in reasonably good performance in terms of the KS distance, AUROC, and FPR, but there are other methods we consider that perform better.

The Reverse Distillation method achieves one of the lower FPRs, but there is not a huge difference between the familiar and unfamiliar score distributions, as evidenced by the lower KS distances and AUROC values. While the maps generated by this method somewhat align with our expectations, they are not particularly intuitive (see Figure \ref{fig:regional-maps}).

The score distributions of FastFlow and PaDiM appear similar to those of Reverse Distillation (see Figure \ref{fig:regional-compare}), but both of these methods achieve higher KS distances, higher AUROC values, and lower FPRs. PaDiM performs better overall, achieving the highest AUROC values across all methods considered. The competency maps of these two methods also generally align with our expectations.

Finally, our method achieves the lowest FPRs, the highest KS distance between familiar and unfamiliar pixels in OOD images, the second highest KS distance between pixels in ID images and unfamiliar pixels in OOD images, and the second highest AUROC values. Our method arguably does the best overall in terms of quantitative metrics, although its performance is not significantly better than that of PaDiM. Our method is also quite intuitive, generating probabilistic scores between zero and one, while the competing method, PaDiM, generates scores between zero and thirty.

\subsection{ADDITIONAL ANALYSIS OF NAVIGATION} \label{app:nav}

Recall from Section \ref{sec:eval} that we conduct ten trials in each of the five navigation scenarios shown in Figure \ref{fig:eval-scenarios}. The results of these experiments are averaged across the five navigation scenarios and summarized in Table \ref{table:navigation}. Here, we present the results for each of the five scenarios individually, which allows for a nuanced evaluation of navigation performance.

Looking at the success rate across navigation scenarios, we notice that the four planning schemes that incorporate regional competency awareness are successful in most situations. However, the turning-based schemes are highly ineffective for the fifth scenario, in which the robot is tasked with navigating between two sets of ladders. This demonstrates that while simply turning in response to low competency regions is effective in many scenarios, there are certain situations for which this response is insufficient. This demonstrated in our demo video\footnoteref{video}. Furthermore, navigation times for trajectory-based schemes tend to be lower than those of turning-based ones, but this difference is very scenario-dependent. Again, the differences in navigation times are the most significant for the fifth navigation scenario. The same idea is generally true regarding path length.

Another observation that is not clear when looking at the results averaged across scenarios is that the baseline controller with no competency awareness generally has the lowest path lengths in successful trials, indicating that it tends to take the most direct path towards the goal. This is intuitive as the baseline controller does not make attempts to avoid low-competency regions but is not captured well in Table \ref{table:navigation} because we only average path lengths across successful trials, and the baseline controller is the most successful in scenario (3), which requires the longest paths to the goal.


\newpage
\begin{table}[h!]
\centering
\vspace{2mm}
\scalebox{0.75}{
\begin{tabular}{|cc|ccccc|}
\hline
Competency & Controller & Success  & Timeout  & Collision  & Navigation  & Path  \\
Awareness & Response &  Rate &  Rate &  Rate &  Time ($s$) &  Length ($m$)  \\
\hline
None & N/A & 90\% & 10\% & 80\% & 57.26 & \textbf{29.46} \\
Overall & Turning & 20\% & 80\% & \textbf{0\%} & 71.37 & 32.17 \\
Regional & Turning & \textbf{100\%} & \textbf{0\%} & \textbf{0\%} & 55.92 & 31.51 \\
Regional & Trajectory & \textbf{100\%} & \textbf{0\%} & 10\% & 55.45 & 29.89 \\
Both & Turning & \textbf{100\%} & \textbf{0\%} & \textbf{0\%} & 56.60 & 31.40 \\
Both & Trajectory & \textbf{100\%} & \textbf{0\%} & \textbf{0\%} & \textbf{54.24} & 29.86 \\
\hline
\end{tabular}
}
\caption{Navigation results for scenario (1): Drive from one side of astronaut to the other.}
\label{table:astro1}
\end{table}

\begin{table}[h!]
\centering
\vspace{2mm}
\scalebox{0.75}{
\begin{tabular}{|cc|ccccc|}
\hline
Competency & Controller & Success  & Timeout  & Collision  & Navigation  & Path  \\
Awareness & Response &  Rate &  Rate &  Rate &  Time ($s$) &  Length ($m$)  \\
\hline
None & N/A & 50\% & 50\% & 80\% & 76.63 & \textbf{33.17} \\
Overall & Turning & 30\% & 70\% & \textbf{0\%} & 87.92 & 42.21 \\
Regional & Turning & 80\% & 20\% & 10\% & 78.01 & 36.70 \\
Regional & Trajectory & \textbf{100\%} & \textbf{0\%} & \textbf{0\%} & \textbf{68.52} & 35.20 \\
Both & Turning & \textbf{100\%} & \textbf{0\%} & \textbf{0\%} & 72.51 & 37.35 \\
Both & Trajectory & 80\% & 20\% & \textbf{0\%} & 75.60 & 36.02 \\
\hline
\end{tabular}
}
\caption{Navigation results for scenario (2): Drive from back to front of astronaut.}
\label{table:astro2}
\end{table}

\begin{table}[h!]
\centering
\vspace{2mm}
\scalebox{0.75}{
\begin{tabular}{|cc|ccccc|}
\hline
Competency & Controller & Success  & Timeout  & Collision  & Navigation  & Path  \\
Awareness & Response &  Rate &  Rate &  Rate &  Time ($s$) &  Length ($m$)  \\
\hline
None & N/A & \textbf{100\%} & \textbf{0\%} & \textbf{0\%} & \textbf{71.90} & \textbf{46.86} \\
Overall & Turning & 0\% & 100\% & \textbf{0\%} & 90.01 & -- \\
Regional & Turning & \textbf{100\%} & \textbf{0\%} & \textbf{0\%} & 84.06 & 48.46 \\
Regional & Trajectory & 90\% & 10\% & \textbf{0\%} & 78.65 & 47.99 \\
Both & Turning & 90\% & 10\% & \textbf{0\%} & 80.98 & 47.37 \\
Both & Trajectory & \textbf{100\%} & \textbf{0\%} & \textbf{0\%} & 75.65 & 47.38 \\
\hline
\end{tabular}
}
\caption{Navigation results for scenario (3): Drive up to front of habitat.}
\label{table:habitat1}
\end{table}

\begin{table}[h!]
\centering
\vspace{2mm}
\scalebox{0.75}{
\begin{tabular}{|cc|ccccc|}
\hline
Competency & Controller & Success  & Timeout  & Collision  & Navigation  & Path  \\
Awareness & Response &  Rate &  Rate &  Rate &  Time ($s$) &  Length ($m$)  \\
\hline
None & N/A & 20\% & 80\% & 90\% & 68.29 & \textbf{23.87} \\
Overall & Turning & 0\% & 100\% & \textbf{0\%} & 75.01 & -- \\
Regional & Turning & \textbf{90\%} & \textbf{10\%} & \textbf{10\%} & 59.78 & 32.25 \\
Regional & Trajectory & \textbf{90\%} & \textbf{10\%} & 30\% & 57.73 & 29.38 \\
Both & Turning & \textbf{90\%} & \textbf{10\%} & \textbf{10\%} & 57.15 & 28.43 \\
Both & Trajectory & \textbf{90\%} & \textbf{10\%} & 50\% & \textbf{54.77} & 27.88 \\
\hline
\end{tabular}
}
\caption{Navigation results for scenario (4): Navigate around both sets of ladders.}
\label{table:ladder1}
\end{table}

\begin{table}[h!]
\centering
\vspace{2mm}
\scalebox{0.75}{
\begin{tabular}{|cc|ccccc|}
\hline
Competency & Controller & Success  & Timeout  & Collision  & Navigation  & Path  \\
Awareness & Response &  Rate &  Rate &  Rate &  Time ($s$) &  Length ($m$)  \\
\hline
None & N/A & 50\% & 50\% & 100\% & 73.98 & \textbf{29.63} \\
Overall & Turning & 0\% & 100\% & 20\% & 90.01 & -- \\
Regional & Turning & 50\% & 50\% & \textbf{20\%} & 83.55 & 38.01 \\
Regional & Trajectory & 80\% & 20\% & 40\% & 73.48 & 33.02 \\
Both & Turning & 0\% & 100\% & 40\% & 90.02 & -- \\
Both & Trajectory & \textbf{90\%} & \textbf{10\%} & \textbf{20\%} & \textbf{63.61} & 31.26 \\
\hline
\end{tabular}
}
\caption{Navigation results for scenario (5): Navigate between two sets of ladders.}
\label{table:ladder2}
\end{table}

\newpage
\clearpage

\addtolength{\textheight}{-12cm}   







\end{document}